\def\year{2022}\relax
\documentclass[letterpaper]{article} 
\usepackage{aaai22}  
\pdfoutput=1
\usepackage{times}  
\usepackage{helvet}  
\usepackage{courier}  
\usepackage[hyphens]{url}  
\usepackage{graphicx} 
\usepackage{natbib}  
\usepackage{caption} 
\DeclareCaptionStyle{ruled}{labelfont=normalfont,labelsep=colon,strut=off} 
\usepackage{algorithm}
\usepackage{algorithmic}
\usepackage{graphicx}
\usepackage{amsmath}
\usepackage{amsthm}

\usepackage{subfigure}
\usepackage{makecell}
\usepackage{multirow}
\usepackage{amssymb}
\usepackage{wrapfig}
\usepackage{newfloat}
\usepackage{listings}
\floatstyle{ruled}
\newfloat{listing}{tb}{lst}{}
\floatname{listing}{Listing}
\title{Unauthorized AI cannot Recognize Me: Reversible Adversarial Example}
\author {
    Jiayang Liu\textsuperscript{\rm 1,}\textsuperscript{\rm 3},
    Weiming Zhang \textsuperscript{\rm 1},
    Kazuto Fukuchi \textsuperscript{\rm 2,}\textsuperscript{\rm 3},
    Youhei Akimoto \textsuperscript{\rm 2,}\textsuperscript{\rm 3},
    Jun Sakuma \textsuperscript{\rm 2,}\textsuperscript{\rm 3}
}
\affiliations {
    \textsuperscript{\rm 1} University of Science and Technology of China\\
    \textsuperscript{\rm 2} University of Tsukuba\\
    \textsuperscript{\rm 3} RIKEN AIP\\
}

\usepackage{bibentry}

\begin{document}

\maketitle

\begin{abstract}
In this study, we propose a new methodology to control how user’s data is recognized and used by AI via exploiting the properties of adversarial examples. For this purpose, we propose reversible adversarial example (RAE), a new type of adversarial example. A remarkable feature of RAE is that the image can be correctly recognized and used by the AI model specified by the user because the authorized AI can recover the original image from the RAE exactly by eliminating adversarial perturbation. On the other hand, other unauthorized AI models cannot recognize it correctly because it functions as an adversarial example. Moreover, RAE can be considered as one type of encryption to computer vision since reversibility guarantees the decryption. To realize RAE, we combine three technologies, adversarial example, reversible data hiding for exact recovery of adversarial perturbation, and encryption for selective control of AIs who can remove adversarial perturbation. Experimental results show that the proposed method can achieve comparable attack ability with the corresponding adversarial attack method and similar visual quality with the original image, including white-box attacks and black-box attacks. 
\end{abstract}

\section{Introduction}
Online services, including social networking services that utilize artificial intelligence trained with users' data, make our lives more convenient.
For example, Facebook uses facial recognition technology to determine whether users are in the photo image data being uploaded and suggest tags indicating the user identities.
Facial recognition technology, including face authentication, is convenient and impactful, but at the same time, it has many ethical issues. There have been concerns pointed out about privacy issues. For example, it would automatically identify an social network software (SNS) user from a selfie uploaded to her anonymous SNS account whose identity is not disclosed. These concerns are not limited to facial images. 
User's gender, race, and other sensitive attributes, including health conditions and diseases, could be estimated from voice data automatically and unintentionally \cite{oh2019speech2face}.
There is no widespread agreement among users on how to handle privacy issues related to data collected from users and the results of recognizing these data through AI (e.g., identities and other sensitive attributes). Technologies that allow users to control how AI recognizes their data would be helpful from the perspective of enhancing privacy protection.

In \cite{gafni2019live}, techniques were proposed to eliminate features in face images of movies that allow identifying the person from face movies without making a significant change in its appearance.
These techniques could, for example, allow users to leave anonymous, natural-looking video messages in public places, which would not be recognized by face recognition unintentionally.
Furthermore, they can prevent users from being tracked for their activities by facial recognition or from being identified from videos uploaded to social networking sites.
Adversarial examples have long been studied as an attack technique to manipulate the identification results of an image without changing its appearance, which can be designed not only for face images \cite{sharif2016accessorize,dong2019efficient,deb2019advfaces}, but also for any objects \cite{xu2020adversarial,eykholt2018robust,cao2019adversarial}. Adversarial example is used initially to make the classification performance of the classifier robust \cite{xie2020adversarial,miyato2018virtual}, while several studies proposed to use adversarial examples to hinder unintentional face image identification from protecting privacy \cite{gafni2019live,zhang2020adversarial}.

The technologies mentioned above allow users to leave anonymous, natural-looking image or video messages in public places without being identified unintentionally. However, one limitation of these technologies is that once features necessary for recognition are eliminated from the data or adversarial perturbation is inserted into data, no one except the holder of the original image can recover the original image. For example, \cite{yangtowards} proposed face encryption by generating adversarial identity masks to protect users' face images from unauthorized face recognition systems. However, this type of face encryption can only be considered as face perturbation since there is no decryption process for it.
Considering that the basic principle of privacy protection is to be able to control one's own information, it is desirable that users themselves can decide which of their images will be recognized and used by AI and which will not. Moreover, we try to not affect users' experience when users share the protected images on social media, and simultaneously conceal their information from unauthorized AI models.

Based on the above considerations, we propose a new methodology to exploit the properties of adversarial examples to control how user’s data is recognized and used by AI. 
For this purpose, we propose reversible adversarial example (RAE), a new type of adversarial example. On the one hand, RAE is able to attack unauthorized AI models to prevent unauthorized access to users' image data since RAE still functions as an adversarial example. On the other hand, authorized AI models can exactly recover the original image from RAE for recognition and further usage.
We explain the functionalities of RAE by taking face recognition system as an example, but the idea of RAE works with any classification task.
Suppose we have a face image $\boldsymbol{X}$ (say, Alice’s face image). Using $\boldsymbol{X}$, we generate a RAE $\boldsymbol{X}’$ with a classification model $\sf f$, which is misrecognized as some label. Then,
\begin{itemize}
\item the appearance of RAE $\boldsymbol{X}’$ is quite similar to original image $\boldsymbol{X}$ to human,
\item $\boldsymbol{X}’$ works as an adversarial example for $\sf f$ (e.g., $\boldsymbol{X}’$ is recognized as Bob),
\item $\boldsymbol{X}’$ is expected to work as an adversarial example for classifiers other than $\sf f$ due to transferability of adversarial examples, and
\item some classifier $\sf f’$ specified by the one who created the RAE (say authorized classifier) can recover $\boldsymbol{X}$ from $\boldsymbol{X}’$ only, meaning only $\sf f’$ can recognize $\boldsymbol{X}’$ correctly as Alice and use the original image $\boldsymbol{X}$.  
\end{itemize}
In other words, RAE can be considered as encryption to computer vision. The reversibility of RAE guarantees the decryption of this type of encryption.

To realize RAE, we employ (1) reversible data hiding \cite{ni2006reversible}, which allows to embed secret information into an image without being able to detected, and (2) highly functional encryption (e.g., attribute-based encryption), which enables to generate ciphertexts in a way that the ones who can decrypt the ciphertexts can be selectively controlled by the private key holder.
 
We demonstrate by experiments on the ImageNet \cite{deng2009imagenet} dataset that RAEs can be misrecognized by unauthorized classifiers, while only the authorized classifier can recover the original images exactly. We demonstrate that the visual quality of RAEs: RAEs look overall similar to original images.

\section{Preliminaries}

\subsection{Adversarial Example}



First proposed by Szegedy \emph{et al.} \cite{szegedy2014intriguing}, various methods to generate adversarial examples have been presented. In this part, we outline the details of the adopted attack methods in this paper. We use $l(\boldsymbol{x},y)$ to notate the cross-entropy loss function where $\boldsymbol{X}$ is the input image and $y$ is the true class for the input image. 

\textbf{FGSM} \cite{goodfellow2014explaining} crafts an adversarial example under the $\ell_{\infty}$ norm as
\begin{equation}
\boldsymbol{X}^{\mathrm{adv}}=\boldsymbol{X}+\epsilon \cdot \operatorname{sign}\left(\nabla_{\boldsymbol{X}} l(\boldsymbol{X}, y)\right)
\end{equation}
where $\nabla_{\boldsymbol{X}} l(\boldsymbol{X}, y)$ denotes the gradient of the loss function with respect to the input image.
 

\textbf{BIM} \cite{kurakin2016adversarialmachine} extends FGSM by taking iterative gradient updates in the following equation: 
\begin{equation}
\boldsymbol{X}^{(t+1), \mathrm{adv}}=\operatorname{clip}_{\boldsymbol{X}, \epsilon}\left(\boldsymbol{X}^{(t),\mathrm{adv}}+\eta \cdot \operatorname{sign}\left(\nabla_{\boldsymbol{X}} l\left(\boldsymbol{X}^{(t),\mathrm{adv}}, y\right)\right)\right)
\end{equation}

where $\operatorname{clip}_{\boldsymbol{X}, \epsilon}$ guarantees the adversarial example to satisfy the $\ell_{\infty}$ constraint. 





\textbf{C\&W} adopts the original C\&W loss \cite{Carlini2017Towards} based on the iterative mechanism of BIM to perform attack in classification tasks. In particular, the loss takes the form of 
\begin{equation}
\boldsymbol{X}^{\mathrm{adv}}= \mbox{argmin}_{\boldsymbol{X}'}  \|\boldsymbol{X}' - \boldsymbol{X}   \| \mbox{ subject to }  Z\left(\boldsymbol{X}'\right)_{y} <  \max_{i\ne y}Z\left(\boldsymbol{X}\right)_{i} - \kappa
\end{equation}
where $Z\left(\boldsymbol{X}\right)_i$ is the logit output of the classifier for the $i$th class.


\subsection{Reversible Data Hiding}


Data hiding \cite{zeng1998digital} is referred to as a method to hide secret information into an image for covert communication. 
Reversible data hiding (RDH) \cite{ni2006reversible} is a special type of data hiding, which allows to recover the original image without any distortion from the marked image and extract the embedded hidden data. Classical RDH algorithms are mainly divided into three categories, compression-embedding\cite{fridrich2002lossless}, difference expansion \cite{tian2003reversible}, and histogram modification \cite{ni2006reversible}. We adopt the histogram modification based RDH in this paper. Intuitively, the algorithm of RDH leverages the fact that the color histograms of natural images distribute unevenly. By shifting the bins of the color histogram in a way that the visual appearance of the image does not change significantly, we can encode a certain amount of information into the image without being detected. 


Let $\boldsymbol{X} \in R^{H \times W \times C}$ be a $C$-channel image (referred to as a cover image) where $H$ and $W$ denotes the height and width of the image, respectively. Let $M \in  \{0,1\}^*$ be a bit-string message of arbitrary length to be encoded into the cover image.  
The RDH process consists of two functions: encoding and decoding. 
We use $\mbox{RDH}:R^{H \times W \times C} \times \{0,1\}^* \rightarrow R^{H \times W \times C}$ to denote the encoding method, which takes an image and a secret bit-string as inputs and outputs an encoded image (referred to as a marked image) with the same size.
$\mbox{RDH}^{-1}:R^{H \times W \times C} \rightarrow R^{H \times W \times C}  \times \{0,1\}^*$ denotes the decoding method, which takes an encoded image as input, and outputs the exact recovery of the original image and the secret bit-string.

\begin{figure}[tbp]
\centering
\includegraphics[width=0.7\linewidth]{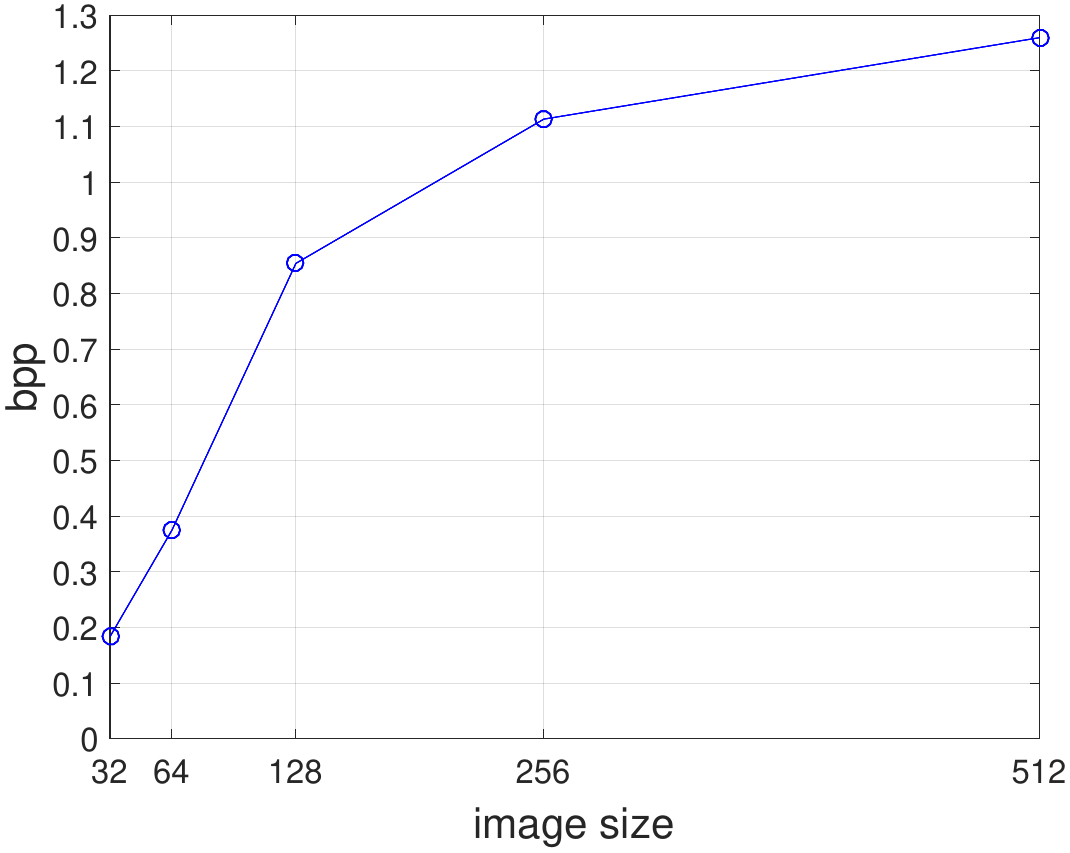}
\caption{ Capacity of RDH with different image sizes.}
\label{bpp}
\end{figure}

For a given integer $a$\footnote{In the process of recovery, $a$ is a key parameter for recovery. So a small part of pixels in the cover image are preserved to embed the value of $a$ utilizing the data hiding technique. Moreover, the information of recovering these preserved pixel values also needs to be treated as a secret message and embedded into the other part of pixels in the cover image utilizing RDH.}, the encoding procedure in one color channel is processed as follows. 
For $i=1, \hdots, H$ and $j=1, \hdots, W$
\begin{equation}
\boldsymbol{X}^{\mbox{RDH}}_{i, j}=\left\{\begin{array}{ll}
\boldsymbol{X}_{i, j}-1, & \text { if } \boldsymbol{X}_{i, j}<a \\
\boldsymbol{X}_{i, j}-M_{(i-1)W+j}, & \text { if } \boldsymbol{X}_{i, j}=a \\
\boldsymbol{X}_{i, j}, & \text { if } \boldsymbol{X}_{i, j}>a
\end{array}\right.
\end{equation}
where $(i,j)$ represents the pixel coordinate in the cover image and $X_{i, j}$ denotes the pixel value at the specified index. The length of binary string $M$ is usually set as shorter than $H \times W \times C$ in RDH (see Figure \ref{bpp} for the detail).
When $a$ is set as the peak of the color histogram, this process shifts the bins less than $a$ towards smaller by 1 to create a vacant bin for data embedding.
Each secret bit in $M$ is embedded into the pixel whose value is $a$ from left to right, top to bottom in the cover image. By taking $a$ as the histogram peak, the capacity of embedding is known to be maximized. As a result, each bit of the string $M$ is embedded into the pixel whose value is $a$ of the cover image and the marked image $\boldsymbol{X}^{\mbox{RDH}}$ is generated as $ \boldsymbol{X}^{\mbox{RDH}}=\mbox{RDH}(\boldsymbol{X},M)$.


On the decoder side, extraction of embedded bits $M$ and recovery of the cover image $\boldsymbol{X}$ is processed as follows:  
\begin{equation}
M=\left\{\begin{array}{ll}
M \parallel 0, & \text { if } \boldsymbol{X}^{\mbox{RDH}}_{i, j}=a \\
M \parallel 1, & \text { if } \boldsymbol{X}^{\mbox{RDH}}_{i, j}=a-1 \\
M, & \mbox{otherwise}
\end{array} \right. 
\end{equation}
\begin{equation}
\boldsymbol{X}_{i, j}=\left\{\begin{array}{ll}
\boldsymbol{X}^{\mbox{RDH}}_{i, j}+1, & \text { if }\boldsymbol{X}^{\mbox{RDH}}_{i, j} \leq a-1 \\
\boldsymbol{X}^{\mbox{RDH}}_{i, j} & \mbox{otherwise}
\end{array}\right.
\end{equation}
where $\parallel $ denotes concatenation. 
We denote these operations by $[\boldsymbol{X}, M]=\mbox{RDH}^{-1}(\boldsymbol{X}^{\mbox{RDH}}).$


\subsubsection{Capacity of RDH}


We test the capacity of RDH on ImageNet \cite{deng2009imagenet}. We adopt the state-of-the-art RDH method \cite{zhang2013recursive}. \cite{zhang2013recursive} is a histogram modification method for RDH which embeds the data by recursively utilizing the decompression and compression processes of an entropy coder. We separately use 1000 images with 3 color channels randomly selected from ImageNet to perform the experiments. We resize the images of ImageNet into different sizes: $32 \times 32 \times 3$, $64 \times 64 \times 3$, $128 \times 128 \times 3$, $256 \times 256 \times 3$ and $512 \times 512 \times 3$. For each image, we gradually increase the length of to-be-embedded bit-string until RDH cannot embed any bit into the image. Then we obtain the maximum length of a bit string that RDH can embed on each image and calculate the embedding rate. Finally, we average the embedding rate of 1000 images to measure the capacity of RDH. Figure \ref{bpp} illustrates the capacity of RDH with different image sizes. The x-axis is the image size and the y-axis is the average bits per pixel (bpp) to be embedded. The result shows that around 1 bit per pixel can be embedded into an image on average. For example, when the size of an image is $299 \times 299 \times 3$ in ImageNet, about 270,000 bits can be embedded covertly. 






\subsection{Highly Functional Encryption}

RDH allows us to embed secret bit-string into an image covertly; however, it does not necessarily mean that the security is guaranteed. We need to use cryptography to guarantee the security of the embedded message. More specifically, we encrypt the secret bits before embedding it into the cover image. Also, we can compress the bit-string before encryption when the size of the bit-string exceeds the capacity of RDH of the cover image. We can use any algorithm for encryption and compression. Unless necessarily mentioned, we suppose the bit-string is compressed and encrypted in the following.

Although any cryptosystem works with RDH, we can introduce highly functional cryptosystem in order to selectively control the AI to recognize the image.
Suppose a communication of secret information between a sender and receiver. Public key cryptosystem is a cryptosystem that uses a pair of public keys $\sf pk$ (which may be known to others, including the sender) and private keys $\sf sk$ (which should be never known to anyone except the receiver). 

In message transmission, anyone can encrypt a message $m \in \{0,1\}^*$ using the receiver's public key as $c \leftarrow \mbox{Enc}_{\sf pk}(m)$ where $c$ is the resulting ciphertext, but the ciphertext can only be decrypted with the receiver's private key as $m \leftarrow \mbox{Dec}_{\sf sk}(c)$.

When a sender wants to specify multiple receivers that can decrypt ciphertexts, attribute-based encryption can be used. Attribute-based encryption, such as \cite{bethencourt2007ciphertext,okamoto2010fully}, enables to encrypt and distribute a message so that the corresponding ciphertext can be decrypted only by entities that satisfy a policy specified at the timing of encryption (e.g., public sector only, authorized services only). Attribute-based encryption enables enforcement of policies by attributes in a non-interactive manner. 
Also, we can use timed-release encryption scheme \cite{dent2007revisiting, matsuda2010efficient} in which the sender specifies the time at which the receiver can decrypt the ciphertext.

In the following sections, we employ RSA instead of a highly functional encryption for simplicity, however this can be exchanged with any kind of high functional encryption without making any modifications.


%
%
%
%
%
%




\section{Proposed Method}

\subsection{Overall Framework}

In our problem setting, we have three stakeholders: user, authorized classifier and unauthorized classifier. The abstract process proceeds as Algorithm \ref{algotithm1}\footnote{Since $I$ is encoded as a bit-string, we can use any lossless information compression algorithm to reduce the size of information to embed. In our experiments, we use arithmetic coder \cite{howard1994arithmetic} for compression.}.

\begin{algorithm}[htb]
	\caption{Reversible Adversarial Example Creation}
    \label{algotithm1}
     {\bf Input: }image $\boldsymbol{X}$ \\
    {\bf Output: }reversible adversarial example $\boldsymbol{X}^{\mbox{RDH}}$\\
	\begin{algorithmic}[1]

        \STATE All classifiers independently create a key pair of public-key cryptosystem and every classifier distributes its public key.
        \STATE User chooses an authorized classifier $\sf f$ and obtains $\sf f$'s public key $\sf pk_f$.
        \STATE User generates an adversarial example $\boldsymbol{X}'$ of image $\boldsymbol{X}$.  
        \STATE User encrypts adversarial perturbation $\Delta \boldsymbol{U} =\boldsymbol{X}' - \boldsymbol{X}$ and auxiliary information $R$ necessary for recovery with the $\sf f$'s public key as $I= \mbox{Enc}_{\sf pk_f}(\Delta \boldsymbol{U}, R)$. Here we simply use RSA algorithm as the encryption method to achieve $\mbox{Enc}_{\sf pk_f}(\Delta \boldsymbol{U}, R)$.
        \STATE User embeds $I$ into the adversarial example $\boldsymbol{X}'$ and obtains RAE as $\boldsymbol{X}^{\mbox{RDH}} = \mbox{RDH}(\boldsymbol{X}', I)$.
	\end{algorithmic}
\end{algorithm}


How the RAE is recognized is summarized as follows:

\begin{itemize}
    \item When a human see $\boldsymbol{X}^{\mbox{RDH}}$, it appears quite similar to $\boldsymbol{X}$ since RDH preserves visual quality of the cover image as demonstrated later. 
    \item When an unauthorized classifier obtains $\boldsymbol{X}^{\mbox{RDH}}$ and recognizes $\boldsymbol{X}^{\mbox{RDH}}$ with his unauthorized classifier, $\boldsymbol{X}^{\mbox{RDH}}$ is expected to be misclassified because $\boldsymbol{X}^{\mbox{RDH}}$ works as an adversarial example. He might be able to obtain $I$ by the decoding process of RDH, however, he can learn nothing from $I$ because $I$ is encrypted by a cryptosystem. So even when the unauthorized classifier learns $I$, exact recovery cannot be obtained. 
\item When an authorized classifier obtains $\boldsymbol{X}^{\mbox{RDH}}$, he obtains $I$ and can obtain $\Delta \boldsymbol{U}, R = \mbox{Dec}_{\sf sk_f}(I)$. With adversarial perturbation $\Delta \boldsymbol{U}$ and auxiliary information $R$, the authorized classifier can obtain the exact recovery $\boldsymbol{X}$ as $\boldsymbol{X} = \boldsymbol{X}^{\mbox{RDH}} - \Delta \boldsymbol{U}$.
\end{itemize}

In the following, we show the detailed realization of each step.

\subsection{Generation of Adversarial Perturbation for RDH}


A direct idea to achieve RAE is to embed adversarial perturbation into the adversarial image utilizing RDH scheme so that the receiver can invalidate the adversarial perturbation. However, RDH is only well suited to embedding a short length of information into a large image. In RDH, Figure \ref{bpp} shows that the average embedding rate is around 1 bit per pixel. Since the size of adversarial perturbation is usually the same size as the cover image, the capacity of RDH is usually not enough to directly embed the adversarial perturbation into the image. For example, if we set the $\ell_{\infty}$ norm perturbation of BIM as 2, the required embedding rate is $\log_2 4 = 2$ bits per pixel. In order to solve this problem, we propose to divide the images into super-pixels (e.g., treat pixels in $2\times2$ tile as a single pixel by smoothing) and then embed adversarial perturbation generated for the super-pixels. 

We denote $\boldmath{X}$ as the original image with size $H \times W \times C$ and $\boldmath{X}^{\prime}$ as its adversarial example with size $H \times W \times C$. Each pixel can take an integer value in $\{0,1, \hdots, 255 \}$.
For each color channel, $C$ ($C$ is either of $r$, $g$ or $b$), the original image $X$ and the adversarial image ${X}^{\prime}$ are divided into non-overlapping tiles with the same size $h \times w$, which are called super-pixels. 
When $h=2$ and $w=2$, the super-pixel consists of a set of four neighboring pixels. 
Let $P_{i,j}$ and $P'_{i,j}$ be the $(i,j)$th super-pixel of the original image and corresponding adversarial example where $1 \leqslant i \leqslant \lceil H/h \rceil, 1 \leqslant j \leqslant \lceil W/w \rceil$.
We consider a type of adversarial perturbation where the pixel values of adversarial perturbation are smoothed over each super-pixel. Due to smoothing, the amount of information contained in such adversarial perturbation can be reduced up to $\frac{1}{h \times w}$, which is sufficiently small to embed with RDH.

\subsubsection{Post Smoothing Method}


This post smoothing super-pixel adversarial attack method is the most straightforward way to realize adversarial example over super-pixels. 
We first generate an adversarial example in an arbitrary way.
Letting $n=h \times w$, we denote the collection of the super-pixels of the image and corresponding adversarial example by $P_{i,j}=\left\{p_{1}, p_{2}, \ldots, p_{n}\right\}$ and $P_{i,j}^{\prime}=\left\{p_{1}^{\prime}, p_{2}^{\prime}, \ldots, p_{n}^{\prime}\right\}$, respectively.
First, calculate the average value of adversarial perturbations of all pixels in each super-pixel and round it to get the closest integer $\Delta \boldsymbol{U}_{i,j}$
\begin{equation}
\Delta \boldsymbol{U}_{i,j}=\operatorname{round}\left(\frac{1}{n} \sum_{k=1}^{n} p_{k}-\frac{1}{n} \sum_{k=1}^{n} p_{k}^{\prime}\right)
.\label{equation1}
\end{equation} 
We call this smoothing method as post smoothing.

Then we get the adversarial example generated with the post smoothing method as $p_{k}^{\prime \prime}=p_{k}+\Delta \boldsymbol{U}_{i,j}$. 
Note that the pixel value $p_{k}^{\prime \prime}$ should be an integer between 0 and 255, so the transformation may result in some overflow/underflow pixel values. For exact recovery, the truncation information for each super-pixel is recorded as $R_{i,j}=\left\{r_{1}, r_{2}, \ldots, r_{n}\right\}$ where $n=h \times w$.





After transformation and truncation, we get a new tile $P_{i,j}^{\prime \prime}$. We get a super-pixel adversarial image $X^{\prime \prime}$ by completing transformations and truncations for all the tiles $P_{i,j}^{\prime \prime}$ where $1 \leqslant i \leqslant \lceil H/h \rceil$ and $1 \leqslant j \leqslant \lceil W/w \rceil$.


\subsubsection{In-the-loop Smoothing Method}

One drawback of the post-smoothing method is that the adversarial perturbation is smoothed after finishing optimization, which can lower the attack success rate. To avoid this, we introduce the in-the-loop smoothing super-pixel adversarial attack method, which is more costly, but is expected to have less effect to the attack success rate. 
To describe how to generate the super-pixel adversarial perturbation, we take BIM \cite{kurakin2016adversarialmachine} as example and propose the in-the-loop smoothing version of BIM.


BIM generates adversarial perturbation by repeatedly update $\boldsymbol{X}$ by the gradient of the loss function with respect to $\boldsymbol{X}$ so that the image is recognized as a wrong label. The idea of the in-the-loop smoothing is to take the gradient with respect to a noise vector $\Delta \boldsymbol{U}$ whose length is the super-pixel number. We initialize the noise vector $\Delta \boldsymbol{U}$ by randomly sampling with uniform distribution in $[-\epsilon / 2, \epsilon / 2]$.
\begin{equation}
\Delta \boldsymbol{U}^{(t)}=\Delta \boldsymbol{U}^{(t-1)}+\eta \cdot \operatorname{sign}\left(\nabla_{\Delta \boldsymbol{U}} l\left(\boldsymbol{X}^{{(t-1)},\mathrm{adv}}, y\right)\right) \label{loop}
\end{equation}

After update, we obtain the $t$th update $\boldsymbol{X}^{(t), \mathrm{adv}}$ by
\begin{equation}
\boldsymbol{X}^{(t), \mathrm{adv}}=\operatorname{clip}_{\boldsymbol{X}, \epsilon}\left(\boldsymbol{X}+f_{pad}(\Delta \boldsymbol{U}^{(t)}, h, w)\right)\label{fmap}
\end{equation}
where function $f_{pad}$ is a mapping function which fills each element of the noise vector $\Delta \boldsymbol{U}^{(t)}$ into a $h \times w$ super-pixel to get the full-size image of adversarial perturbation padded with the super-pixel values.
Note that $\operatorname{clip}_{\boldsymbol{X}, \epsilon}$ operates on super-pixels of $\boldsymbol{X}$ and guarantees that the perturbations within each super-pixel are still same. 
A super-pixel adversarial example is generated by alternate iterations of eq. \ref{loop} and eq. \ref{fmap}. 

\begin{figure}[tbp]
\centering
\subfigure[$2 \times 2$]{
\begin{minipage}[b]{0.23\linewidth}
\includegraphics[width=0.92\linewidth]{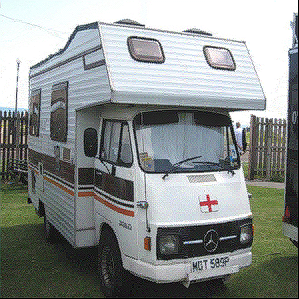}\
\includegraphics[width=1\linewidth]{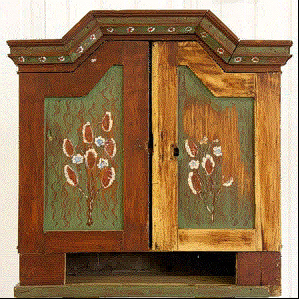}\
\includegraphics[width=0.92\linewidth]{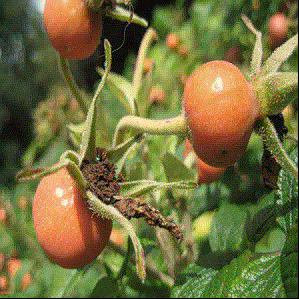}\
\includegraphics[width=0.92\linewidth]{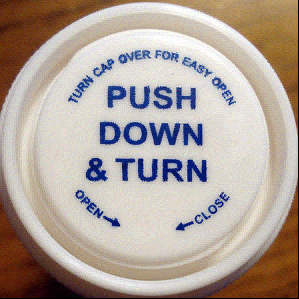}
\end{minipage}}
\subfigure[$3 \times 3$]{
\begin{minipage}[b]{0.23\linewidth}
\includegraphics[width=0.92\linewidth]{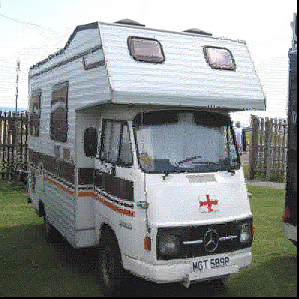}\
\includegraphics[width=1\linewidth]{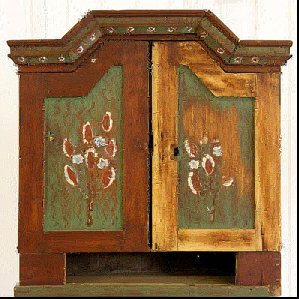}\
\includegraphics[width=0.92\linewidth]{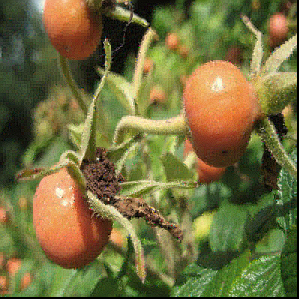}\
\includegraphics[width=0.92\linewidth]{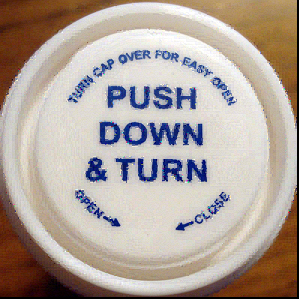}
\end{minipage}}
\subfigure[$4 \times 4$]{
\begin{minipage}[b]{0.23\linewidth}
\includegraphics[width=0.92\linewidth]{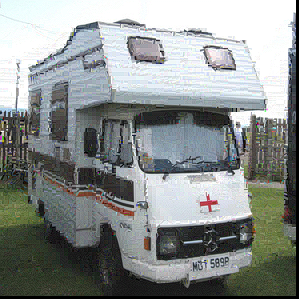}\
\includegraphics[width=1\linewidth]{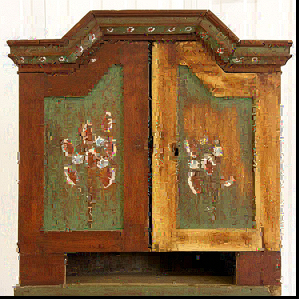}\
\includegraphics[width=0.92\linewidth]{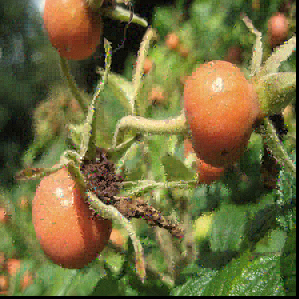}\
\includegraphics[width=0.92\linewidth]{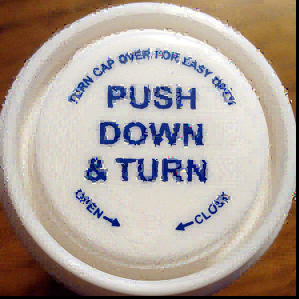}
\end{minipage}}
\subfigure[$5 \times 5$]{
\begin{minipage}[b]{0.23\linewidth}
\includegraphics[width=0.92\linewidth]{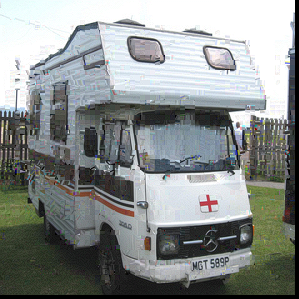}\
\includegraphics[width=1\linewidth]{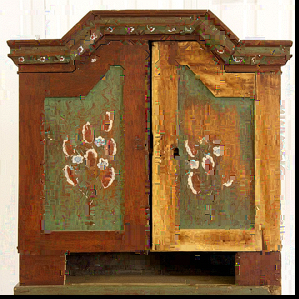}\
\includegraphics[width=0.92\linewidth]{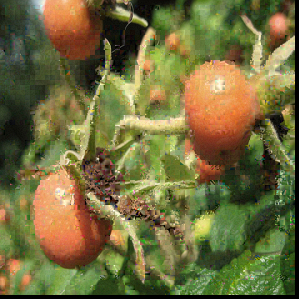}\
\includegraphics[width=0.92\linewidth]{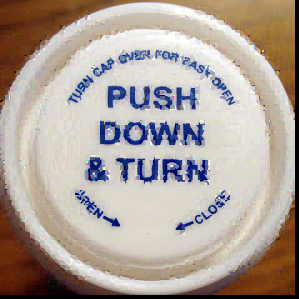}
\end{minipage}}
\caption{Visual quality of reversible adversarial examples with different super-pixel sizes.}
\label{block}
\end{figure}

\section{Experiments}


In this section, we experimentally evaluate to what extent the RAEs can deceive the DNN. If RAEs have a similar attack success rate to regular adversarial examples, we can say unauthorized classifiers cannot recognize given images correctly.

Thanks to RDH, the scheme of the proposed method guarantees that the authorized classifier can exactly recover the original image from the RAE without any distortion. So we do not evaluate the classification accuracy of adversarial from the viewpoint of the authorized classifier. We confirmed that RAEs generated for our experiments were recovered to be the original images exactly.

\begin{table}[tbp]
\caption{The attack success rate (\%) of adversarial examples (AE) and reversible adversarial examples (RAE) on the ImageNet dataset in the white-box setting (Inception-v3).}
\begin{center}

\begin{tabular}{c|c|c|c}
 \hline
 Attack Method & Parameter & AE & RAE \\
 \hline
 \hline
 \multirow{3}{*}{FGSM}  & $\epsilon=4$ &76.70  &71.58 \\
\cline{2-4}
& $\epsilon=8$ &83.82 &79.87 \\
\cline{2-4}
& $\epsilon=16$ &88.18 &86.08 \\
\hline
 \multirow{3}{*}{BIM (post)}  & $\epsilon=4$ &96.71  &73.84\\
 \cline{2-4}
& $\epsilon=8$ &97.83 &83.61\\
\cline{2-4}
& $\epsilon=16$ &99.78 &89.45 \\
 \hline
 \multirow{3}{*}{BIM (in-the-loop)}  & $\epsilon=4$ &96.71 &81.01\\
 \cline{2-4}
& $\epsilon=8$ &97.83 &94.67\\
\cline{2-4}
& $\epsilon=16$ &99.78 &97.41 \\
 \hline
 \multirow{3}{*}{C\&W (post)}  & $\kappa=0$ &98.62  &28.68 \\
 \cline{2-4}
& $\kappa=50$ &99.20 &81.43 \\
 \cline{2-4}
& $\kappa=100$ &99.68 &95.45 \\
 \hline
\end{tabular}

\end{center}
\label{CW}
\end{table}

\begin{table*}[tbp]
\caption{The attack success rate (\%) of adversarial examples and reversible adversarial examples on the ImageNet dataset. $*$ indicates the target model.}
\centering
\begin{tabular}{l|c|c|c|c}
\hline
Model          &white-box  &\multicolumn{3}{c}{black-box} \\ \hline 
Attack Method          &Inc-v3*  &Inc-v4  &IncRes-v2 &Inc-v3$_{adv}$\\ \hline \hline
FGSM	         & 83.82    & 53.25   & 55.90   & 55.08   \\ 
RAE-FGSM	 & 79.87    & 48.28   & 50.54   & 50.12 \\ \hline
BIM	         & 97.83    & 40.72   & 37.23      & 30.82    \\ 
RAE-BIM (post)	 & 83.61    & 27.34   & 24.37       & 19.67    \\ 
RAE-BIM (in-the-loop)	 & 94.67    & 35.31   & 30.73       & 25.57    \\ \hline
C\&W	         & 99.68    & 2.33   & 1.75      & 0.94    \\ 
RAE-C\&W (post)	 & 95.45    & 1.12   & 0.85      & 0.34    \\ \hline

\end{tabular}
\label{FGSM and BIM}
\end{table*}

We perform our experiments on 100,000 images randomly selected from the ImageNet \cite{deng2009imagenet}. Since only correctly classified images are considered for evaluation of the attack ability, the classification accuracy of the original set is 100\%. The pretrained Inception-v3 \cite{szegedy2016rethinking} is adopted as the default target attack model, which is evaluated with top-1 accuracy. The values of pixels per color channel of the images range from 0 to 255. The adversarial attack methods used in the experiments are FGSM, BIM, and C\&W, and we perform untargeted attacks in the experiments. For FGSM, we set the perturbation budget $\epsilon$ as 4/255,  8/255 and 16/255. For BIM, we set the perturbation budget $\epsilon$ as 4/255, 8/255 and 16/255, the number of iterations as 20, the step size $\eta$ as 1/255. We use $\ell_{\infty}$ norm distance metric for FGSM and BIM. For C\&W, we use ${l_2}$ distance metric and $\kappa = 0, 50, 100$ where $\kappa$ is the parameter of attack confidence.


\subsection{Selection of Super-pixel Size}

Intuitively, smaller super-pixel size leads to less influence on the attack success rate of the generated RAE while larger super-pixel size leads to larger amount of data can be embedded using RDH. In addition, we illustrate several samples of RAEs with different super-pixel sizes. To have a clearer difference, in-the-loop smoothing version of BIM with $\epsilon = 4 $ is used here. As shown in Figure \ref{block}, smaller super-pixel size achieves better visual quality of RAEs. So we choose smaller super-pixel size in the experiments to achieve higher attack success rate and better visual quality of RAEs.

For FGSM, the required embedding rate for RDH is $\log_2 2 = 1$ bits per pixel. This embedding rate is within the capacity of RDH. So we set the super-pixel size of FGSM as $1 \times 1$. That is to say, we use RDH to directly embed the adversarial perturbation of FGSM into the adversarial image to generate the RAE. So the in-the-loop smoothing method for FGSM turns to be same as the post smoothing method for FGSM when super-pixel size of FGSM is $1 \times 1$. 

For BIM, the required embedding rate of $\epsilon=4$ is $\log_2 8 = 3$ bits per pixel, the required embedding rate of $\epsilon=8$ is $\log_2 16 = 4$ bits per pixel and the required embedding rate of $\epsilon=16$ is $\log_2 32 = 5$ bits per pixel. So we set the super-pixel size of BIM with $\epsilon=4$ as $1 \times 3$, $\epsilon=8$ as $2 \times 2$ and $\epsilon=16$ as $1 \times 5$. Here we use the post smoothing method and the in-the-loop smoothing method for BIM.

For C\&W, the bit length of adversarial perturbation for different images changes a lot. So we gradually increase the super-pixel size for each adversarial image from $1 \times 1$ to $1 \times 2$, $1 \times 3$ and $2 \times 2$ until the capacity of RDH is enough to embed the perturbation. Here we use the post smoothing method for C\&W because the optimization of the in-the-loop smoothing method did not complete within the practical computation time. We suspect the reason may be that in-the-loop version of C\&W is not so effective in finding the suitable adversarial perturbations for super-pixels.




\subsection{Attack Ability in the White-box Setting}

To evaluate the attack ability, we adopt the attack success rate against the pretrained classification models on ImageNet dataset. Table \ref{CW} shows results on the attack success rate of reversible adversarial examples (RAE) generated by FGSM, BIM and C\&W with parameters of different values in the white-box setting. The column ``AE'' is the result of adversarial examples while ``RAE'' is the result of reversible adversarial examples. We compare three baselines FGSM, BIM and C\&W with our corresponding RAE. For BIM, we presented the results of two variants, post smoothing and in-the-loop smoothing. For FGSM, the in-the-loop smoothing method is same as the post smoothing method when super-pixel size of FGSM is $1 \times 1$. For C\&W, only the result of the post smoothing method is shown. 


From the result, we can see that the gap of the attack success rate  between AE and RAE of FGSM is less than 6 percent. For the in-the-loop smoothing method of BIM, the gap of the attack success rate between AE and RAE is less than 4 percent when the perturbation budget $\epsilon$ is not smaller than 8/255. For C\&W, the gap of the attack success rate between AE and RAE is less than 5 percent when $\kappa$ is set as 100.
So the attack success rates of regular adversarial examples and RAEs in the white-box setting are quite close when we set larger perturbation budgets, indicating RAEs can work correctly. Moreover, the success rate of RAE generated by the in-the-loop smoothing method of BIM is at least 7 percent higher than the post smoothing method of BIM, which reveals that the results of in-the-loop smoothing gives better attack success rate compared to the post smoothing method. This is because adversarial perturbation generated with the in-the-loop smoothing is made considering the super-pixel strategy while that of post smoothing ignores this in the process of adversarial perturbation generation. Finally, we get the conclusion from Table \ref{CW} that RAE generated by larger budget adversarial perturbations can achieve comparable attack ability with regular adversarial examples in the white-box setting.


\subsection{Visual Results}
\begin{figure}[htbp]
\centering
\subfigure{
\begin{minipage}[b]{0.23\linewidth}
\includegraphics[width=0.83\linewidth]{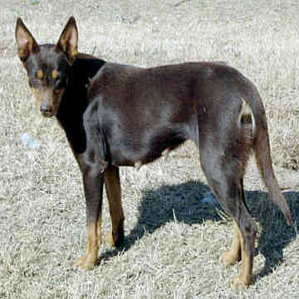}\
\includegraphics[width=0.83\linewidth]{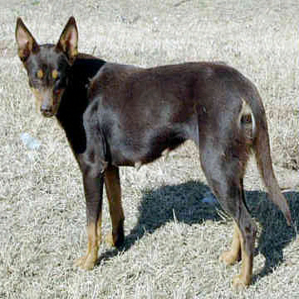}\
\includegraphics[width=0.83\linewidth]{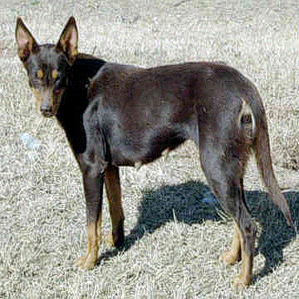}
\end{minipage}}
\subfigure{
\begin{minipage}[b]{0.23\linewidth}
\includegraphics[width=0.83\linewidth]{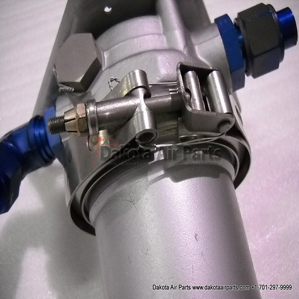}\
\includegraphics[width=0.83\linewidth]{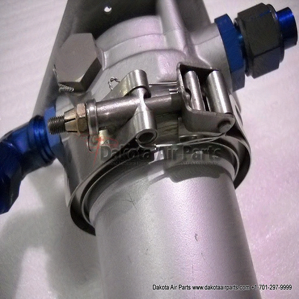}\
\includegraphics[width=0.83\linewidth]{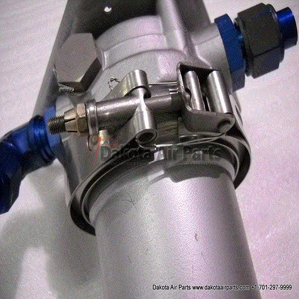}
\end{minipage}}
\subfigure{
\begin{minipage}[b]{0.23\linewidth}
\includegraphics[width=0.83\linewidth]{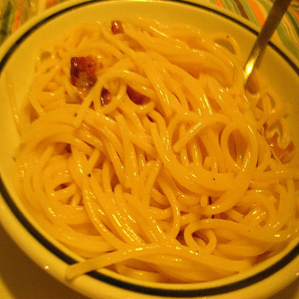}\
\includegraphics[width=0.83\linewidth]{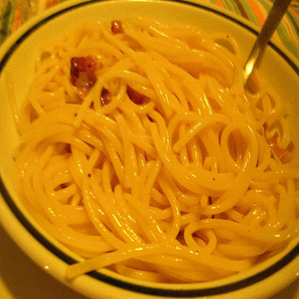}\
\includegraphics[width=0.83\linewidth]{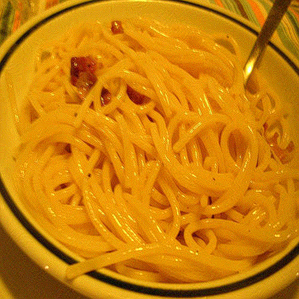}
\end{minipage}}
\subfigure{
\begin{minipage}[b]{0.23\linewidth}
\includegraphics[width=0.83\linewidth]{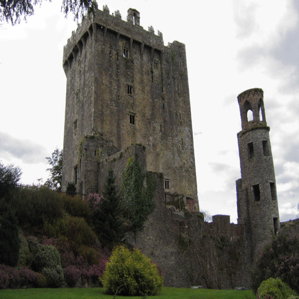}\
\includegraphics[width=0.83\linewidth]{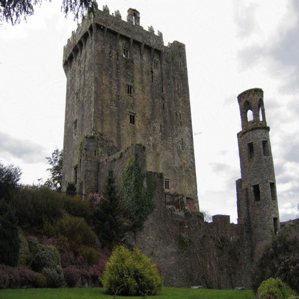}\
\includegraphics[width=0.83\linewidth]{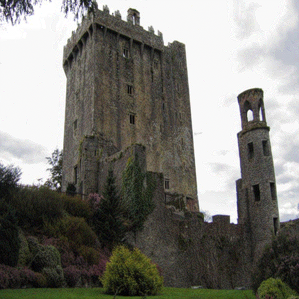}
\end{minipage}}
\caption{Visual results of original images (top), adversarial examples generated by BIM with $\epsilon = 4/255$ (middle) and reversible adversarial examples generated by in-the-loop smoothing method for BIM with $\epsilon = 4/255$ (bottom).}
\label{figure3}
\end{figure}

\begin{figure}[htbp]
\centering
\subfigure{
\begin{minipage}[b]{0.23\linewidth}
\includegraphics[width=0.83\linewidth]{./picture/BIM/clean/111.png}\
\includegraphics[width=0.83\linewidth]{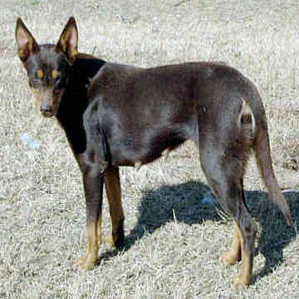}\
\includegraphics[width=0.83\linewidth]{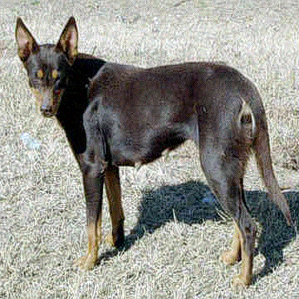}
\end{minipage}}
\subfigure{
\begin{minipage}[b]{0.23\linewidth}
\includegraphics[width=0.83\linewidth]{./picture/BIM/clean/112.png}\
\includegraphics[width=0.83\linewidth]{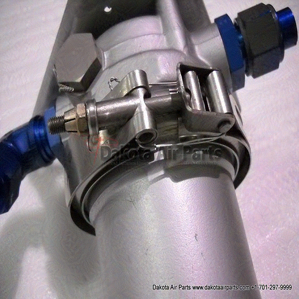}\
\includegraphics[width=0.83\linewidth]{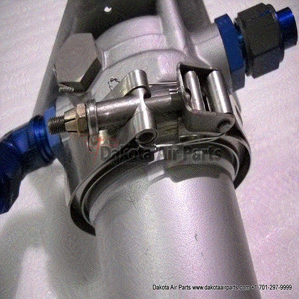}
\end{minipage}}
\subfigure{
\begin{minipage}[b]{0.23\linewidth}
\includegraphics[width=0.83\linewidth]{./picture/BIM/clean/114.png}\
\includegraphics[width=0.83\linewidth]{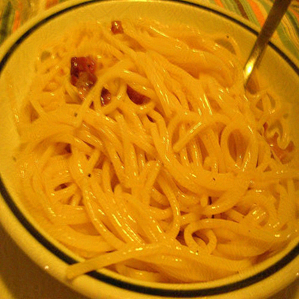}\
\includegraphics[width=0.83\linewidth]{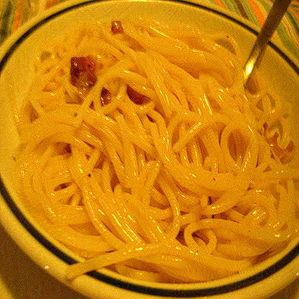}
\end{minipage}}
\subfigure{
\begin{minipage}[b]{0.23\linewidth}
\includegraphics[width=0.83\linewidth]{./picture/BIM/clean/119.png}\
\includegraphics[width=0.83\linewidth]{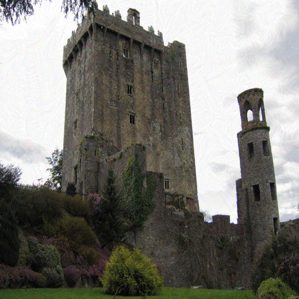}\
\includegraphics[width=0.83\linewidth]{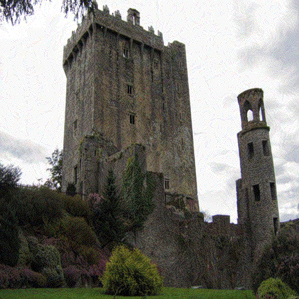}
\end{minipage}}
\caption{Visual results of original images (top), adversarial examples generated by BIM with $\epsilon = 8/255$ (middle) and reversible adversarial examples generated by in-the-loop smoothing method for BIM with $\epsilon = 8/255$ (bottom).}
\label{figure4}
\end{figure}

\begin{figure}[htbp]
\centering
\subfigure{
\begin{minipage}[b]{0.23\linewidth}
\includegraphics[width=0.83\linewidth]{./picture/BIM/clean/111.png}\
\includegraphics[width=0.83\linewidth]{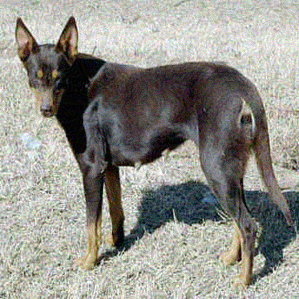}\
\includegraphics[width=0.83\linewidth]{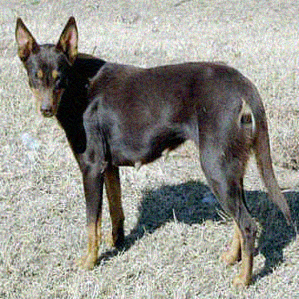}
\end{minipage}}
\subfigure{
\begin{minipage}[b]{0.23\linewidth}
\includegraphics[width=0.83\linewidth]{./picture/BIM/clean/112.png}\
\includegraphics[width=0.83\linewidth]{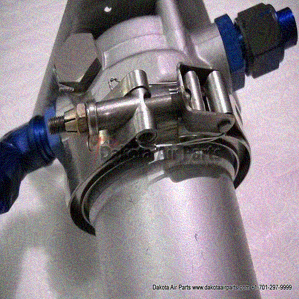}\
\includegraphics[width=0.83\linewidth]{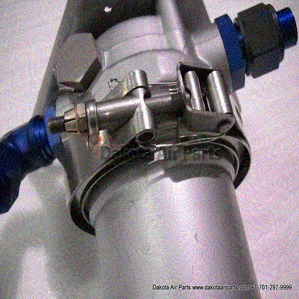}
\end{minipage}}
\subfigure{
\begin{minipage}[b]{0.23\linewidth}
\includegraphics[width=0.83\linewidth]{./picture/BIM/clean/114.png}\
\includegraphics[width=0.83\linewidth]{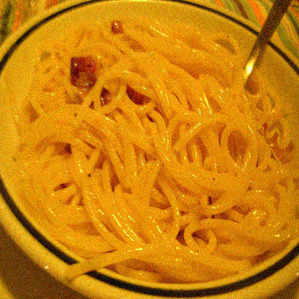}\
\includegraphics[width=0.83\linewidth]{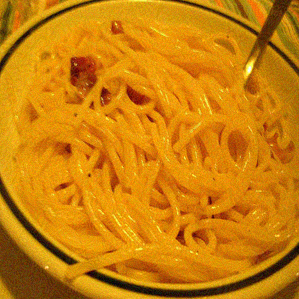}
\end{minipage}}
\subfigure{
\begin{minipage}[b]{0.23\linewidth}
\includegraphics[width=0.83\linewidth]{./picture/BIM/clean/119.png}\
\includegraphics[width=0.83\linewidth]{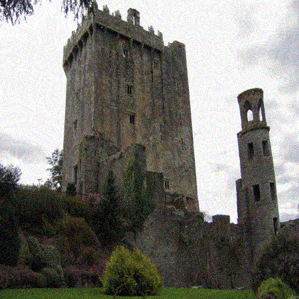}\
\includegraphics[width=0.83\linewidth]{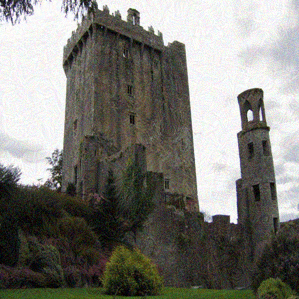}
\end{minipage}}
\caption{Visual results of original images (top), adversarial examples generated by BIM with $\epsilon = 16/255$ (middle) and reversible adversarial examples generated by in-the-loop smoothing method for BIM with $\epsilon = 16/255$ (bottom).}
\label{figure5}
\end{figure}

To evaluate the visual quality of RAE with different perturbation budgets, we show the visual results of in-the-loop smoothing method for BIM with $\epsilon = 4/255$, $\epsilon = 8/255$ and $\epsilon = 16/255$ respectively in Figure \ref{figure3}, \ref{figure4} and \ref{figure5}. For each row, we show the original images (top), the adversarial examples (middle) and the RAEs (bottom) respectively. We can see that the adversarial images and their corresponding RAEs look almost same. When the perturbation budget becomes larger, the visual quality of RAE becomes worse to human eyes. But human eyes can hardly perceive the difference between RAEs and the original images when the perturbation budget $\epsilon$ is not larger than 8. These visual results reveal that RAE can still achieve good visual quality to human eyes when the adversarial perturbations are not large. 


\subsection{Attack Ability in the Black-box Setting}

In real applications, the black-box setting is more important for RAE.
So we also evaluate the attack success rate of RAEs with larger perturbation budget in the black-box setting. The intention of the experiment setting is that users mainly rely on the transferability of RAEs in real applications since users usually can not have direct access to the authorized/unauthorized classifiers.
In Table \ref{FGSM and BIM}, we generated RAEs with Inc-v3, and evaluated the attack success rate with Inception-v4 \cite{szegedy2017inception}, Inception Resnet-v2 \cite{szegedy2017inception}, and Inception-v3$_{adv}$.
Here, Inception-v3$_{adv}$ denotes Inception-v3 \cite{szegedy2016rethinking} model with ensemble adversarial training \cite{tramer2017ensemble} which is known to be robust against adversarial attack.
We set the perturbation budget $\epsilon$ as 8/255 for FGSM and BIM, and the attack confidence $\kappa$ as 100 for C\&W.
We can confirm that attack success rate in the black-box setting is lowered compared to the white-box setting in both regular adversarial examples and RAEs. Also, the gap between regular adversarial examples and RAEs is enlarged. To have better attack success rate in the black-box setting, introducing techniques to increase the transferability \cite{xie2019improving,huang2019enhancing} into RAE is straightforward.  

\section{Discussion}
This paper proposes the first prototype framework of RAE and makes a preliminary attempt in the black-box setting. To have better attack success rate in the black-box setting, introducing techniques to increase the transferability can be a straightforward solution. In addition, reducing the drop of attack ability in black-box setting can also be exploited by improving the reversible data hiding algorithm or using more efficient encryption schemes, such as ``key agreement protocol'' + ``symmetrical encryption'', to reduce the length of information that reversible data hiding needs to embed. 


The purpose of RAE is to control how user's data is recognized by AI. We believe this proposal would have positive societal impact. However, attackers with malicious purpose may use RAE to attack AI. Since RAE is one type of adversarial example, it can have similar potential negative societal impacts as adversarial example. So RAE will also pose a potential threat to safety and security critical applications. To deal with the potential threat of RAE, adversarial defense methods can be adopted as countermeasures to protect deep neural networks.


\section{Conclusion}

In this paper, we propose reversible adversarial example to control how user's data is recognized and used by AI. The AI specified by the user can exactly recover the original image from the RAE while other AIs cannot recognize the RAE correctly. This paper proposes the concept of RAE and give the first prototype framework to verify its feasibility. The proposed method combines adversarial example, reversible data hiding and encryption to realize RAE. Experiments show that the proposed method can achieve comparable attack ability with the corresponding adversarial attack method and similar visual quality with the original image. In the future, we will further investigate how to increase the attack ability of the RAE in the black-box setting.

\bibliography{aaai22}

\end{document}